\documentclass[letterpaper]{article} % DO NOT CHANGE THIS
\usepackage{aaai25}  % DO NOT CHANGE THIS
\usepackage{times}  % DO NOT CHANGE THIS
\usepackage{helvet}  % DO NOT CHANGE THIS
\usepackage{courier}  % DO NOT CHANGE THIS
\usepackage[hyphens]{url}  % DO NOT CHANGE THIS
\usepackage{graphicx} % DO NOT CHANGE THIS
\usepackage{natbib}  % DO NOT CHANGE THIS AND DO NOT ADD ANY OPTIONS TO IT
\usepackage{caption} % DO NOT CHANGE THIS AND DO NOT ADD ANY OPTIONS TO IT
\usepackage{algorithm}
\usepackage{algorithmic}

\usepackage{newfloat}
\usepackage{listings}
\DeclareCaptionStyle{ruled}{labelfont=normalfont,labelsep=colon,strut=off} % DO NOT CHANGE THIS
\floatstyle{ruled}
\newfloat{listing}{tb}{lst}{}
\floatname{listing}{Listing}
\usepackage{amsmath} 
\usepackage{amssymb}
\usepackage{multirow} 
\usepackage{booktabs} 
\usepackage{makecell}
\usepackage{subcaption}

\usepackage{amsfonts}
\usepackage{booktabs}
\usepackage{tabularx}
\usepackage{xcolor}
\usepackage{colortbl}
\usepackage{longtable}

\usepackage{hyperref}
\hypersetup{
    colorlinks=true,        
    citecolor=black,
    linkcolor=black
}

\def \best#1{\textbf{#1}}

\title{QTG-VQA: Question-Type-Guided Architectural for VideoQA Systems}
\author {
    Zhixian He\textsuperscript{\rm },
    Pengcheng Zhao\textsuperscript{\rm },
    Fuwei Zhang\textsuperscript{\rm },
    Shujin Lin\textsuperscript{\rm }\thanks{Corresponding author.}
}
\affiliations {
    \textsuperscript{\rm }Sun Yat-Sen University, Guangzhou, China\\
    \{hezhx29, zhaopch6, zhangfw5\}@mail2.sysu.edu.cn, \\
    linshjin@mail.sysu.edu.cn
}

\begin{document}

\maketitle

\begin{abstract}
In the domain of video question answering (VideoQA), the impact of question types on VQA systems, despite its critical importance, has been relatively under-explored to date. However, the richness of question types directly determines the range of concepts a model needs to learn, thereby affecting the upper limit of its learning capability. This paper focuses on exploring the significance of different question types for VQA systems and their impact on performance, revealing a series of issues such as insufficient learning and model degradation due to uneven distribution of question types. Particularly, considering the significant variation in dependency on temporal information across different question types, and given that the representation of such information coincidentally represents a principal challenge and difficulty for VideoQA as opposed to ImageQA. To address these challenges, we propose QTG-VQA, a novel architecture that incorporates question-type-guided attention and adaptive learning mechanism. Specifically, as to temporal-type questions, we design Masking Frame Modeling technique to enhance temporal modeling, aimed at encouraging the model to grasp richer visual-language relationships and manage more intricate temporal dependencies. Furthermore, a novel evaluation metric tailored to question types is introduced. Experimental results confirm the effectiveness of our approach.
\end{abstract}


\section{Introduction}
In recent years, the rapid development of deep learning has provided substantial technical support for video question answering (VideoQA) research, especially the advancements in vision-language models have enabled researchers to design increasingly complex models capable of processing multimodal information from video content and text \cite{zhou2020unified, 10205457, chen2023tem}. This has led to unprecedented abilities in discovering patterns in data, extracting visual and textual representations. However, prior works have overlooked the importance of high-level overview information such as the type of questions in VQA datasets, inadvertently limiting the models' capabilities.

\begin{figure}[ht]
    \centering
    \includegraphics[width=\linewidth]{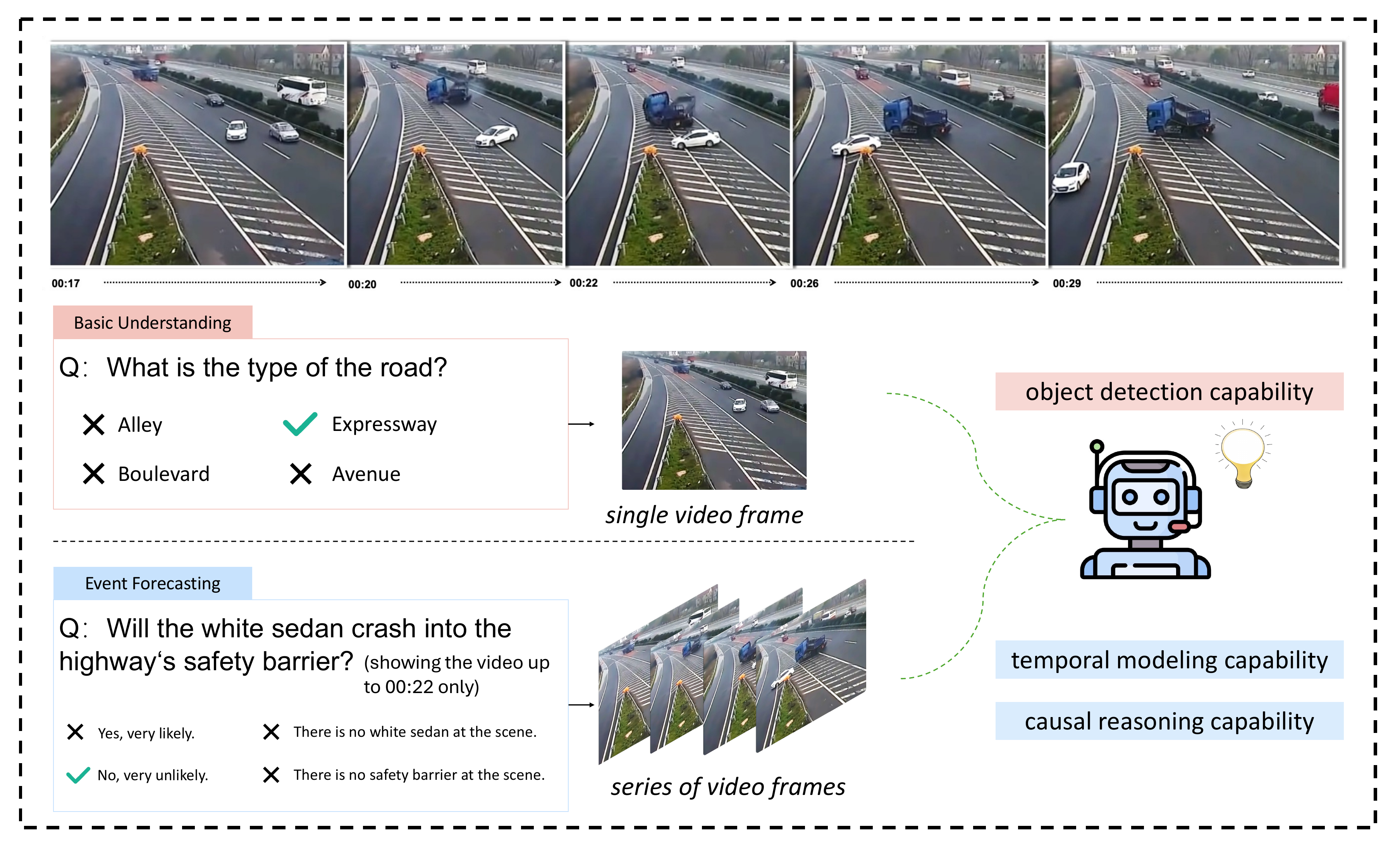}
    \caption{Illustration of different problem types in video question answering. ``\textit{Basic understanding}" and ``\textit{Event Forecasting}" type problems have differing requirements for input sequence dependency and model capabilities.}
    \label{fig:intro}
\end{figure}

Existing research, to a significant extent, utilizes a unified approach for the processing of multimodal information \cite{yang2022frozenbilm, chen2020uniter, li2020oscar, Su2020VL-BERT, bao2022vlmo}, focusing on the interaction between visual and textual information, failing to adequately consider the potential impact of question types on model performance. In video question answering tasks, the diversity of question types not only determines the range of concepts a model needs to learn \cite{zhang2016yin}, thereby influencing its learning capacity \cite{goyal2017making, johnson2016clevr, teney2016zeroshot, hudson2019gqa}, but also indicates how a model should interpret the relationship between visual content and question text \cite{farazi2020attention}. For instance, as the Figure \ref{fig:intro} shows, questions categorized under ``Basic Understanding" may only require the model to possess basic object detection capabilities, whereas ``Event Forecasting" questions demand the model to master understanding time series and perform advanced causal reasoning. Consequently, the diversity and complexity of question types play a crucial role in guiding models to deeply learn the intricate relationships between vision and language.

In light of this, this paper introduces QTG-VQA, an innovative architecture that utilizes question type to guide attention mechanisms and the model training process. Specifically, our work is as follows:

Firstly, we designed a question-type-guided model architecture QTG-VQA, which converts question types into learnable embedding vectors using one-hot encoding. The question type serves as meta-information, providing additional context about the nature of the question. Building upon this, we further implement an Attention-Weighted Multi-Task Adaptive Learning approach (AWMTL), featuring an adaptive learning loop, both the attention weights and learning rates during the model’s learning process based on the distribution and difficulty level of the question types. Such a design allows the model to leverage question types to guide attention allocation and resource distribution during learning.

Secondly, for question types such as ``Event Forecasting" that are heavily reliant on temporal information, we introduce the Masking Frame Modeling technique. This technique is grounded in a Transformer autoregressive model and aims to enhance the model's understanding of long-range temporal dependencies in videos. It includes two predictive tasks: predicting future frames and reconstructing masked frames within a video sequence. The inspiration for this approach is drawn from the `Masked Language Model' pre-training task of the BERT model \cite{Devlin2019BERTPO} in the natural language processing field, which is based on the observation that understanding the context of natural language and parsing the temporal dynamics in videos share similarities. In text processing, the extensive context between words shapes the meaning of sentences. Similarly, in video analysis, the relationships between frames reveal the evolution of actions and clues to the storyline.

Furthermore, considering the significant differences in difficulty and the varying demands on the model's learning capabilities among different question types, traditional methods of evaluating average accuracy fail to reflect the true performance of the model. This paper introduces two novel evaluation metrics: the Inverse Frequency-Weighted Average Accuracy (IFWAA) and the Equal Weight Average Accuracy (EWAA), which are used to assess the model's performance across different question types.

Extensive comparative experiments were conducted on the SUTD-TrafficQA \cite{xu2021sutd} dataset to evaluate the effectiveness of the QTG-VQA model. Additionally, ablation studies were performed to highlight the significance of Question Type Guided Attention and Masking Frame Modeling in achieving performance gains. Overall, our contributions are summarized as follows:
\begin{itemize}
    \item We introduce the QTG-VQA architecture that uniquely leverages question type to guide the model's attention mechanisms and training process, incorporating AWMTL method that dynamically adjusts weights based on question types.
    \item For questions demanding an understanding of temporal dynamics, we propose the Masking Frame Modeling technique, which improves the model's proficiency in recognizing long-range temporal dependencies.
    \item Addressing the shortcomings of traditional evaluation metrics, our paper proposes a novel assessment method based on question types, which includes both IFWAA and EWAA.
\end{itemize}

\section{Related Work}
Video Question Answering (VQA) is a multimodal understanding task that necessitates accurate comprehension and complex reasoning across visual and textual modalities to identify the correct answer for a given video-text pair. This paper primarily focuses on the multiple-choice VideoQA task, aiming to predict the correct answer from multiple provided candidate answers.

\subsection{VideoQA Datasets}
The earliest video question answering dataset is MovieQA \cite{tapaswi2016movieqa}, which was based on 408 movies, compiling a total of 14,944 questions. These questions ranged from simple inquiries such as ``who did what to whom" to more complex queries about ``why" and ``how" certain events occurred. After that, TGIF-QA \cite{jang2017tgif} was constructed, which introduced video-level spatiotemporal reasoning questions for the first time, and categorized them into ``Repetition Count," ``Repeating Action," ``State Transition," and ``Frame QA." Up to the present, researchers have undertaken a series of studies aimed at constructing more sophisticated and comprehensive Q\&A datasets \cite{zhu2017uncovering, fan2019egovqa, Kim2017DeepStoryVS, lei2019tvqa, lei2019tvqaplus, castro-etal-2020-lifeqa, li2020hero, choi2020dramaqa, zadeh2019social, zhao2020video, colas2019tutorialvqa, xu2021sutd, li2024sports}. According to statistics in this paper, the datasets that include question types at the annotations are TGIF-QA \cite{jang2017tgif}, EgoVQA \cite{fan2019egovqa}, PororoQA \cite{Kim2017DeepStoryVS}, VideoQA(FiB) \cite{zhu2017uncovering}, LifeQA \cite{castro-etal-2020-lifeqa}, STAR \cite{wu2024star}, Sports-QA \cite{li2024sports}, etc. Whereas those that do not are TVQA \cite{lei2019tvqa}, TVQA+ \cite{lei2019tvqaplus}, How2QA \cite{li2020hero}, DreamQA \cite{choi2020dramaqa}, Social-IQ \cite{zadeh2019social}, PsTuts-VQA \cite{zhao2020video}, TutorialVQA \cite{colas2019tutorialvqa}, etc. For detailed statistics please refer to Table A.1 in Appendix A. During the organization of datasets, it was discovered that although a majority of researchers consciously categorize questions into different types during data collection, this global information is not utilized during the model training process.

\begin{figure*}[!t]
    \centering
    \includegraphics[width=\textwidth]{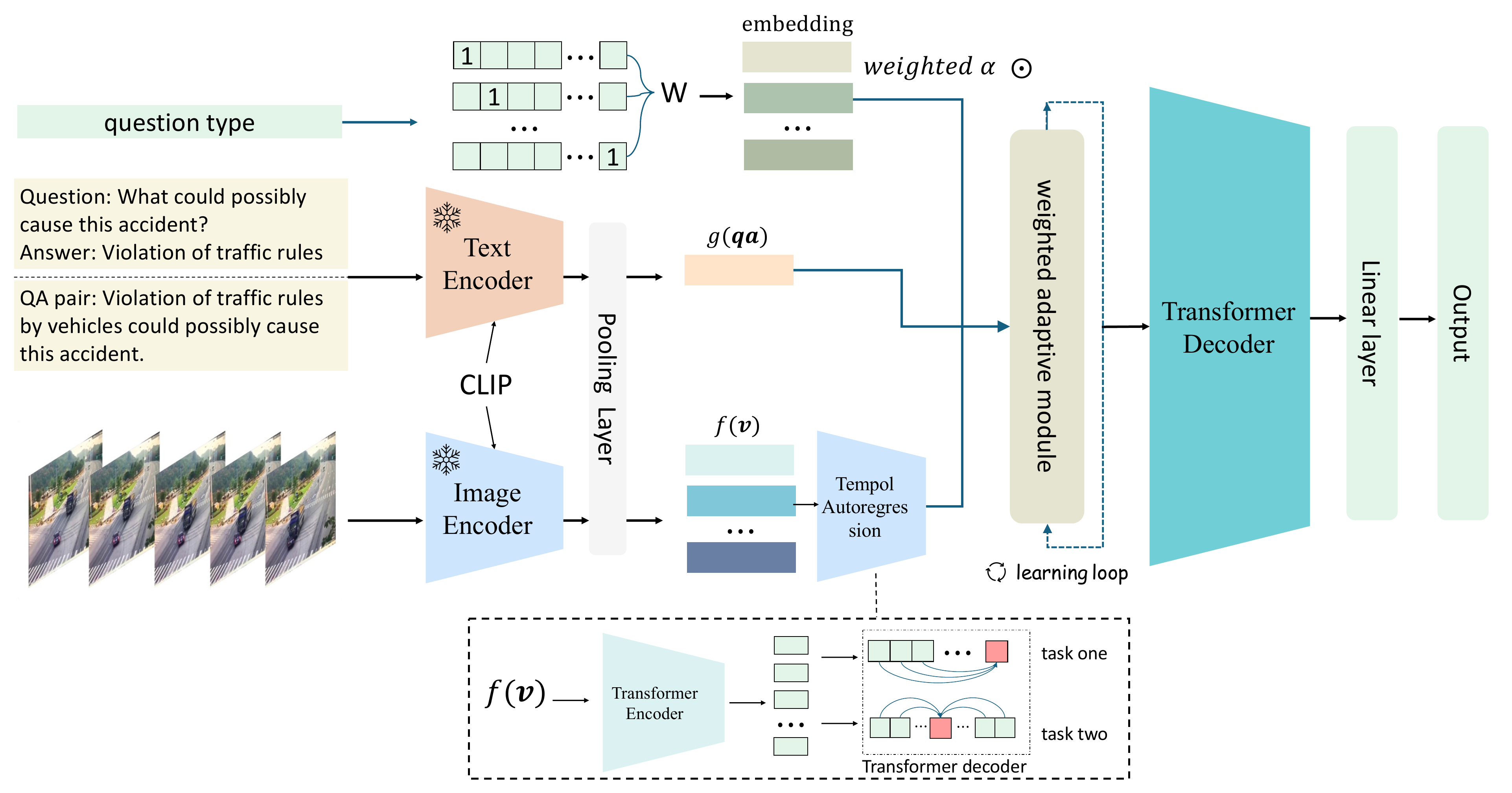}
    \caption{The overall framework of QTG-VQA. The architecture is primarily composed of four core components: visual-text feature extractor, question type embedding module, weighted adaptive module, and temporal autoregression module}
    \label{fig:framework}
\end{figure*}

\subsection{Learning Bias}
A recent survey on video question answering categorized existing Q\&A datasets into Factoid VideoQA and Inference VideoQA based on the technical challenges faced in understanding the questions and reasoning the answers \cite{zhong-etal-2022-video}. Factoid VideoQA focuses on answering questions whose answers can be directly obtained from the video content, such as identifying objects or activities performed by characters at specific times. In contrast, Inference VideoQA involves more complex logical reasoning and interaction across different modalities. The survey highlighted that complex inferential questions are key to enhancing models' understanding capabilities. However, our front review of the datasets revealed that the majority of Q\&A datasets do not strictly differentiate between factoid and inferential questions but rather train in a mixed manner. As previously noted the type of questions can introduce bias \cite{yang2020gives}, we believed this approach leads to models exhibiting artificially high performance on evaluation metrics \cite{kafle2017visual, kafle2017analysis} and a decline in learning capability on complex reasoning questions involving causality and temporality \cite{Kervadec_2021_CVPR}. By monitoring the training and evaluation process for different question types through QTG-VQA, we can not only identify the differential capabilities of models in handling factoid and inferential questions but also pinpoint specific reasoning tasks where the model underperforms.

\section{Method}

\subsection{Overview}
In this study, we introduce the QTG-VQA model architecture, a both flexible and concise framework that can be easily integrated into a variety of existing complex models. Figure \ref{fig:framework} shows an overview of QTG-VQA’s framework. In this section, we first provide an overview of the entire implementation framework, which is rooted in the classical dual-stream question answering model, then followed by a detailed explanation of specific implementation strategies. This includes the integration of question type embeddings, the application of Masking Frame Modeling techniques, the design of loss functions, and the introduction of new evaluation metrics based on question types.

\subsection{Question Type Guided Model Architecture}
This section presents the overall framework of the QTG-VQA, following the practice of most current models that apply vision-language pretraining models to downstream visual question answering tasks \cite{zhong2023stoa, lin2022egocentric, zhang2024vision, dai2024instructblip}. The architecture is primarily composed of four core components: visual-text feature extractor, question type embedding module, weighted adaptive module, and temporal autoregression module.

\textbf{feature extractor and q-type embedding.} In our QTG-VQA model, the visual-text feature extractor employs a frozen CLIP \cite{radford2021learning} model (ViT-B/32) to obtain initial encodings of text and image features. To accommodate the input requirements of CLIP, we follows the video frame sampling technique of previous methods~\cite{xu2021sutd}, uniformly select 8 key frames over a long video sequence, then pick 16 consecutive frames centered around each of these key frames to capture comprehensive temporal contexts. A total of 128 frames are selected for each video. For the extracted image features \( f(v_t) \) and text features \( g(q,a) \), they are first reduced to the same 128-dimensional space through an averaging pooling layer.
\begin{equation}
\bar{f}(v)=\frac{1}{T} \sum_{t=1}^T f\left(v_t\right), \quad \bar{g}(q, a)=\frac{1}{L} \sum_{l=1}^L g(q, a)_l
\end{equation}
where \(T\) is the total number of video frames, and \(L\) is the number of words in the text. To leverage question types to guide the attention mechanism, we first encode the question type into a one-hot encoded vector. Each question type is represented as a vector of length \(N\), where \(N\) is the total number of question types. This one-hot encoding is then transformed into a dense embedding vector through a trainable weight matrix \(W\),
\begin{equation}
e_{q} = W \cdot \text{one-hot}(q_{\text{idx}}), \quad W \in \mathbb{R}^{D \times N}
\end{equation}
where \(W\) is the weight matrix that maps the one-hot encoding into a \(D\)-dimensional embedding space. This embedding vector \(e_{q}\) is then utilized in the Transformer decoder to modulate attention weights, thereby enhancing the model's focus on relevant features for the given question type. 

~\\
\textbf{AWMTL Implementation.} Following the feature extraction and question type embedding, we employ the Attention-Weighted Multi-Task Adaptive Learning (AWMTL) to dynamically adapt the learning process. This approach adjusts the attention weights and learning rates based on the observed performance metrics (loss and accuracy) for each question type, aiming to optimize the model's focus and efficiency in learning from diverse data characteristics. The attention weights are adjusted using a learning feedback loop.
\begin{equation}
P_{q} = \alpha \cdot \text{Loss}_{q} + (1 - \alpha) \cdot (1 - \text{Acc}_{q})
\end{equation}
where \(\text{Loss}_{q}\) and \(\text{Acc}_{q}\) are the loss and accuracy of the model for question type \(q\), and \(\alpha\) is a balancing coefficient. The attention weights are then updated using,
\begin{equation}
w_{q} = \frac{\exp(-\text{P}_{q})}{\sum_{j} \exp(-\text{P}_{j})}
\end{equation}
This softmax function ensures that more attention is paid to question types where the model performs poorly, thereby enhancing the learning focus on more challenging aspects. Furthermore, the learning rates for each question type are also adapted based on their respective attention weights.
\begin{equation}
\eta_{q} = \eta \cdot w_{q}
\end{equation}
where \(\eta\) is the base learning rate. This ensures that learning rates are higher for question types that require more intensive learning, aligning resource allocation with the need for improvement. With the dynamically adapted attention weights and learning rates, the Transformer decoder fuses the features as follows:
\begin{equation}
F = TD(\bar{f}(v), \bar{g}(q,a), e_{q} \cdot w_{q}; \Theta_{\text{D}})
\end{equation}
where \(TD\) represents the Transformer decoder function, \(\Theta_{\text{D}}\) are the parameters of the decoder, and \(e_{q} \cdot w_{q}\) denotes the weighted embedding vectors based on adjusted attention. The resultant feature \(F\) is then projected back into a 512-dimensional feature space using a linear projection layer \(P\), forming the final feature representation:
\begin{equation}
\hat{F} = P(F)
\end{equation}
where \(P\) efficiently consolidates the diverse information into a cohesive feature set suited for downstream tasks.

~\\
\textbf{temporal autoregression module.} To enhance the understanding of temporal dynamics in video sequences, our model incorporates a Temporal Autoregression Module, which performs two key tasks: predicting future frames and reconstructing masked frames. This module is implemented through an autoregressive Transformer decoder, with the mathematical expressions as follows:
\begin{equation}
\begin{split}
    &\hat{f}(v_i) = f_d(f(v_{1:i-1}) \mid f, g; \psi)  \\
    &\hat{f}(v_{\text{masked}}) = f_d(f(v_{\text{context}}) \mid f, g; \psi)
\end{split}
\end{equation}
Here, $f(v_{1:i-1})$ represents the sequence of features from the first to the $(i-1)$-th frame, serving as the input to the decoder, while $f(v_{\text{context}})$ refers to the features of frames surrounding the masked frame, providing adequate context for inference and reconstruction. $f$ and $g$ denote the latent visual embeddings and text embeddings associated with the question type, respectively, and $\psi$ denotes the set of parameters for the decoder.

Through these mechanisms, the Temporal Autoregression Module not only processes existing video information to predict future events but also reconstructs complete video content when some information is intentionally obscured. This approach significantly enhances the model's capability to understand and process the temporal dynamics in video sequences, particularly in video question answering tasks requiring complex temporal reasoning.

\subsection{Loss Function Design}
In this study, we designed a comprehensive loss function to optimize the performance of the multiple-choice video question answering model. This loss function combines the multi-choice loss, the reconstruction loss from the temporal autoregression module, and two question type-based losses, aiming to improve the model's overall performance. The specific loss components are as follows:

\textbf{Multi-Choice Loss.} We use a modified Hinge Loss \cite{jang2017tgifqa} for the multi-choice task, defined as:
\begin{equation}
L_{\text{hinge}} = \max(0, \delta - (s_c - s_w))
\end{equation}
where \( s_c \) is the score of the correct answer, \( s_w \) is the highest score among the incorrect options, and \( \delta \) is a margin.

\textbf{Reconstruction Loss.} We use Mean Squared Error (MSE) loss to assess the temporal autoregression module's performance:
\begin{equation}
L_{\text{mse}} = \frac{1}{T} \sum_{t=1}^{T} \|\hat{f}(v_t) - f(v_t)\|_2^2
\end{equation}
where \( \hat{f}(v_t) \) and \( f(v_t) \) are the predicted and actual frames, respectively.

\textbf{Question Type Losses.} To address different question types, we incorporate:
\begin{equation}
\begin{split}
    &L_{\text{avg\_type}} = \frac{1}{N} \sum_{i=1}^{N} L_{\text{qtype}}^{(i)}  \quad \\
    &L_{\text{freq\_weighted}} = \sum_{i=1}^{N} w_i \cdot L_{\text{qtype}}^{(i)} \quad 
\end{split}
\end{equation}
where \( L_{\text{avg\_type}} \) averages the losses across all question types, and \( L_{\text{freq\_weighted}} \) weights each type's loss \( L_{\text{qtype}}^{(i)} \) by its frequency \( w_i \).

Then the final loss function is:
\begin{equation}
L = L_{\text{hinge}} + L_{\text{mse}} + L_{\text{avg\_type}} + L_{\text{freq\_weighted}}
\end{equation}

\subsection{Type-Specific Cost Evaluation}
In the evaluation of video question answering systems, traditional methods typically use average accuracy as the performance metric \cite{chen2023tem, whitehead2022reliable}. However, this approach has a significant flaw: it does not account for the uneven distribution of question types within the dataset. This can lead the model to bias towards more frequently occurring question types while neglecting less frequent or more challenging ones. To address this issue, we propose two metrics: Inverse Frequency Weighted
Average Accuracy \textbf{(IFWAA)} and Equal Weight Average
Accuracy \textbf{(EWAA)}. The specific formula is as follows:
\begin{equation}
\textbf{IFWAA} = \frac{\sum_{i=1}^{N} \frac{acc_i}{f_i}}{\sum_{i=1}^{N} \frac{1}{f_i}}
\end{equation}

\begin{equation}
\textbf{EWAA} = \frac{1}{N} \sum_{i=1}^{N} acc_i
\end{equation}
where \(acc_i\) is the accuracy for question type \(i\), \(f_i\) is the frequency of question type \(i\) in the dataset, and \(N\) is the total number of question types.

\section{Experiment}
To validate the effectiveness of the proposed QTG-VQA approach and explore the generalization capabilities across different question types, we conducted comprehensive experiments on the SUTD-TrafficQA dataset. Additionally, we provide ablation studies to investigate the impact of each component.

\subsection{Experimental Setup}
\textbf{Datasets.} The SUTD-TrafficQA \cite{xu2021sutd} dataset is dedicated to VideoQA tasks under traffic scenarios, which comprises over 10,000 video clips covering diverse traffic incidents, accompanied by more than 62,535 pairs of manually curated questions and answers. Of these, 56,460 pairs are allocated for training and 6,075 for testing. SUTD-TrafficQA aims to evaluate the cognitive capabilities of video event understanding models in complex traffic contexts, featuring six challenging traffic-related reasoning tasks: ``Basic Understanding," ``Event Forecasting," ``Reverse Reasoning," ``Counterfactual Inference," ``Introspection," and ``Attribute Attribution." Each task is presented in a multiple-choice format with an unrestricted number of possible answers. Examples of each question type are provided in Appendix Table B.1. Furthermore, the sample counts for each question type in the training and validation sets are shown in Figure \ref{fig:intro}. 

\begin{figure}[ht]
    \centering
    \includegraphics[width=\linewidth]{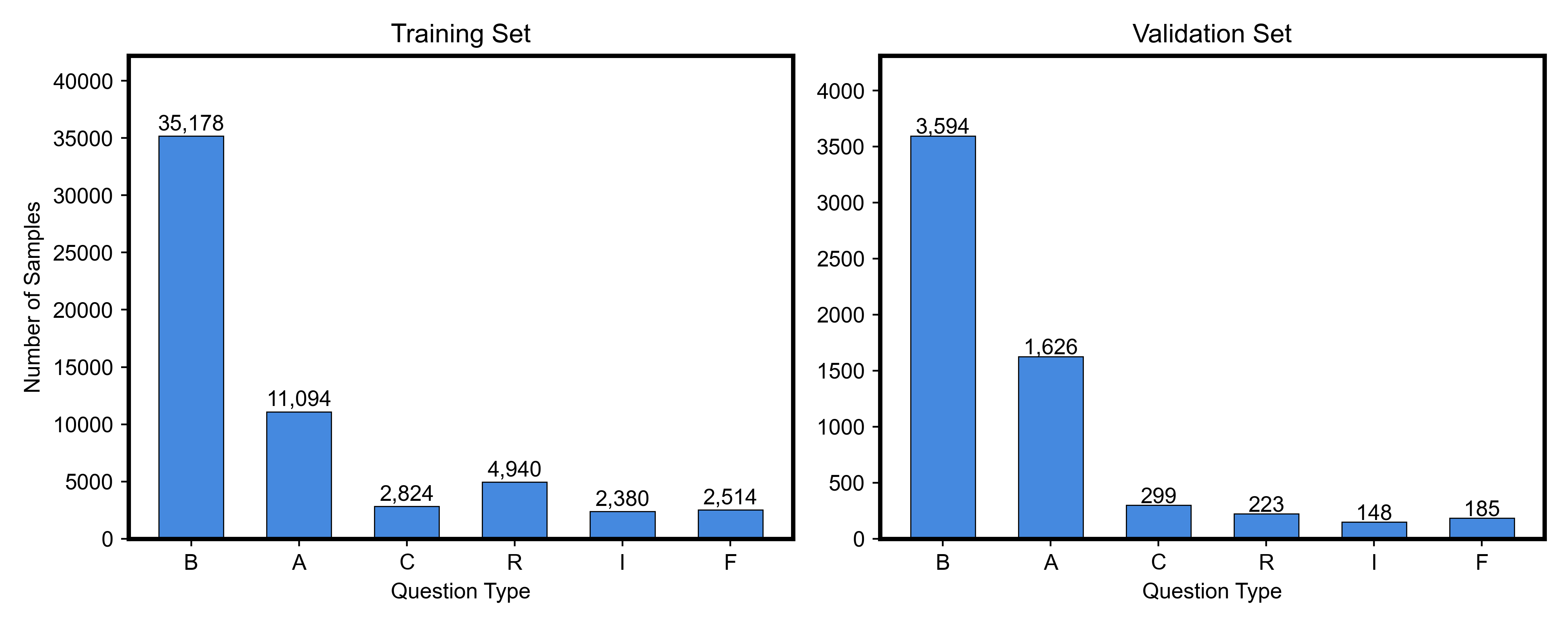}
    \caption{Sample counts across different question types in SUTD-TrafficQA dataset for training and validation sets. The x-axis shows letter abbreviations for each question type.}
    \label{fig:counts}
\end{figure}

\textbf{Implementation Details.} When implementing the QTG-VQA model, we used PyTorch 1.11.0 and employed the pre-trained CLIP (ViT-B/32) \cite{radford2021learning} to extract text and image embeddings. The question type embeddings were mapped to a 512-dimensional space to guide the attention weight adjustments in the Transformer decoder, with the weights normalized using softmax. The decoder was configured with 16 attention heads, 512-dimensional embeddings, and a 2048-dimensional feed-forward layer. We incorporated the AWMTL method to dynamically adjust the learning rates and attention weights based on the loss and accuracy of each question type. The temporal autoregression module shares the same Transformer decoder configuration as the main model, including 512-dimensional input-output embeddings and a 2048-dimensional feed-forward layer. Training was conducted with the Adam optimizer \cite{kingma2014adam}, with an initial learning rate of 1e-4 and a batch size of 128. The final output features were projected into a 512-dimensional space. Detailed hyperparameter settings are provided in the Appendix D.

\textbf{Baseline Models}. To evaluate the performance of our model and currently existing models, we compare QTG-VQA with the other 11 baseline methods including Unsupervised CLIP \cite{radford2021learning}, Unsupervised CLIP + Language template, Totally fine-tuning, Partially fine- tuning, LORA \cite{hu2021lora}, CLIP-Adapter \cite{gao2024clip}, Multi-layer CLIP-Adapter, Prompt learning (change words)(with/without using adapter heads) \cite{zhou2022learning}, Prompt learning (add words) \cite{jia2022visual} and Tem-adapter \cite{chen2023tem}. The detailed descriptions are shown in Appendix C.

\subsection{Main Experiments Results}
As shown in Table \ref{tab:exmain}, our proposed QTG-VQA method, while achieving the highest average accuracy (Avg-acc) of 45.6\%, also scored the highest in the newly defined, question-type-related evaluation metrics—Inverse Frequency Weighted Average Accuracy (IFWAA) and Equal Weight Average Accuracy (EWAA)—with scores of 47.6\% and 47.4\%. For question types, QTG-VQA retained top performance on frequent B-type questions and showed notable improvements in rarer and more challenging types, such as F-type questions, which rose from 38.1\% to 44.8\%, with all types reaching peak performance. These results highlight that QTG-VQA not only excels in handling common question types but also demonstrates superior learning ability and adaptability in challenging question categories, overcoming the issue of model degradation caused by the uneven distribution of question types.

\begin{table*}[!t] 
\centering
\begin{tabular}{@{}lccccccccc@{}} 
%\toprule
\hline \hline
\addlinespace[3pt] 
\multirow{2}{*}{\makecell{\textbf{Methods}}} & \multicolumn{9}{c}{\textbf{SUTD-TrafficQA}} \\
\cmidrule(lr){2-10} 
& \textbf{B} & \textbf{F} & \textbf{R} & \textbf{C} & \textbf{I} & \textbf{A} & \textbf{Avg-acc} & \textbf{IFWAA} & \textbf{EWAA} \\ 
\midrule
Unsupervised CLIP \cite{radford2021learning} & 25.6 & 20.1 & 34.0 & 30.8 & 22.8 & 28.8 & 26.5 & 26.3 & 27.0 \\ 
CLIP + Template \cite{radford2021learning} & 31.8 & 36.0 & 29.9 & 71.8 & 22.1 & 33.4 & 32.3 & 39.7 & 37.5 \\ 
Totally finetuning \cite{Guo_2023_CVPR} & 39.8 & 35.1 & 46.6 & 45.6 & 37.2 & 40.5 & 40.3 & 40.6 & 40.8 \\ 
Partially finetuning \cite{Guo_2023_CVPR} & 41.6 & 37.8 & 44.6 & 50.0 & 33.1 & 41.7 & 41.7 & 40.7 & 41.5 \\ 
LORA \cite{hu2021lora} & 38.7 & 38.7 & 36.7 &	37.9 & 34.5 & 38.1 & 38.3 &	36.8 & 37.4 \\
CLIP-Adapter \cite{gao2024clip} & 35.8 & 32.0 & 35.4 & 42.3 & 33.1 & 32.1 & 34.8 & 35.6 & 35.1 \\ 
Multi-layer Adapter \cite{pan2022stadapter} & 30.5 & 26.6 & 26.5 & 38.5 & 28.3 & 25.8 & 29.1 & 30.2 & 29.4 \\ 
Prompt learning (C) \cite{zhou2022learning} & 42.4 & 32.4 & 45.2 & 55.5 & 40.7 & 43.6 & 42.9 & 43.6 & 43.3 \\ 
Prompt learning (C*) \cite{zhou2022learning} & 40.3 & 33.2 & 41.0 & 46.5 & 34.9 & 38.4 & 39.7 & 38.7 & 39.3 \\ 
Prompt learning (A) \cite{jia2022visual} & 41.7 & 31.5 & 40.1 & 48.4 & 33.1 & 41.4 & 41.1 & 38.2 & 39.4 \\ 
Tem-adapter \cite{chen2023tem} & 44.5 & 38.1 & 45.5 & 51.4 & 36.5 & 47.0 & 45.0 & 43.8 & 42.6 \\ 
\midrule
QTG-VQA (Ours) & \textbf{44.7} & \textbf{44.8} & \textbf{48.2} & \textbf{56.8} & \textbf{41.9} & \textbf{47.8} & \textbf{45.6} & \textbf{47.6} & \textbf{47.4} \\ 
\hline \hline
%\bottomrule
\end{tabular}
\caption{Performance comparison on SUTD-TrafficQA dataset.}
\label{tab:exmain}
\end{table*}

\subsection{Question-Type Guided Attention Ablation}
\begin{figure}[!t]
\centering
\begin{subfigure}[b]{0.5\textwidth}
    \centering
    \includegraphics[width=\textwidth]{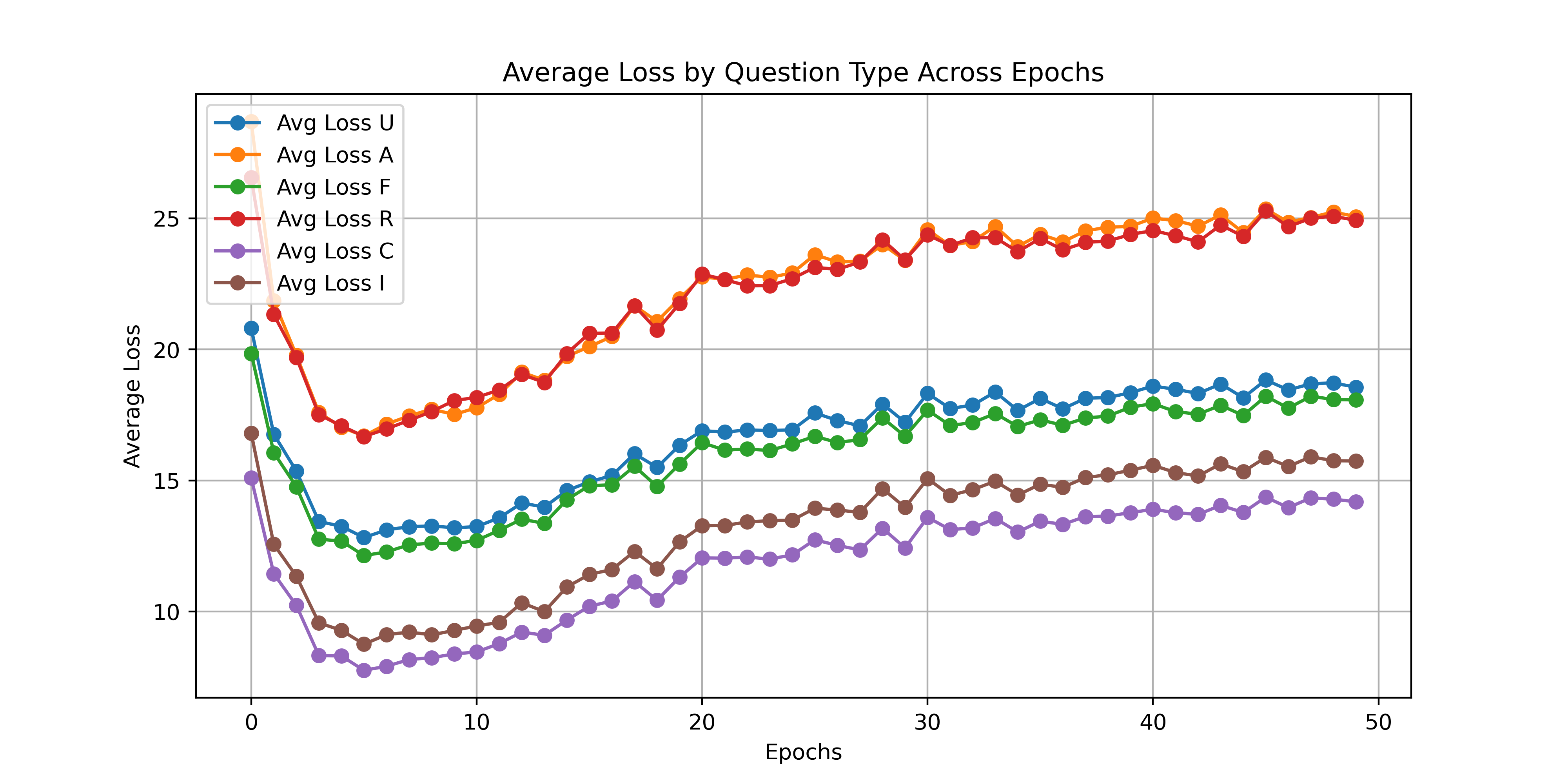}
    %\caption{}
    \label{fig:before_qtype_loss}
\end{subfigure}
\vspace{-7mm} 
\begin{subfigure}[b]{0.5\textwidth}
    \centering
    \includegraphics[width=\textwidth]{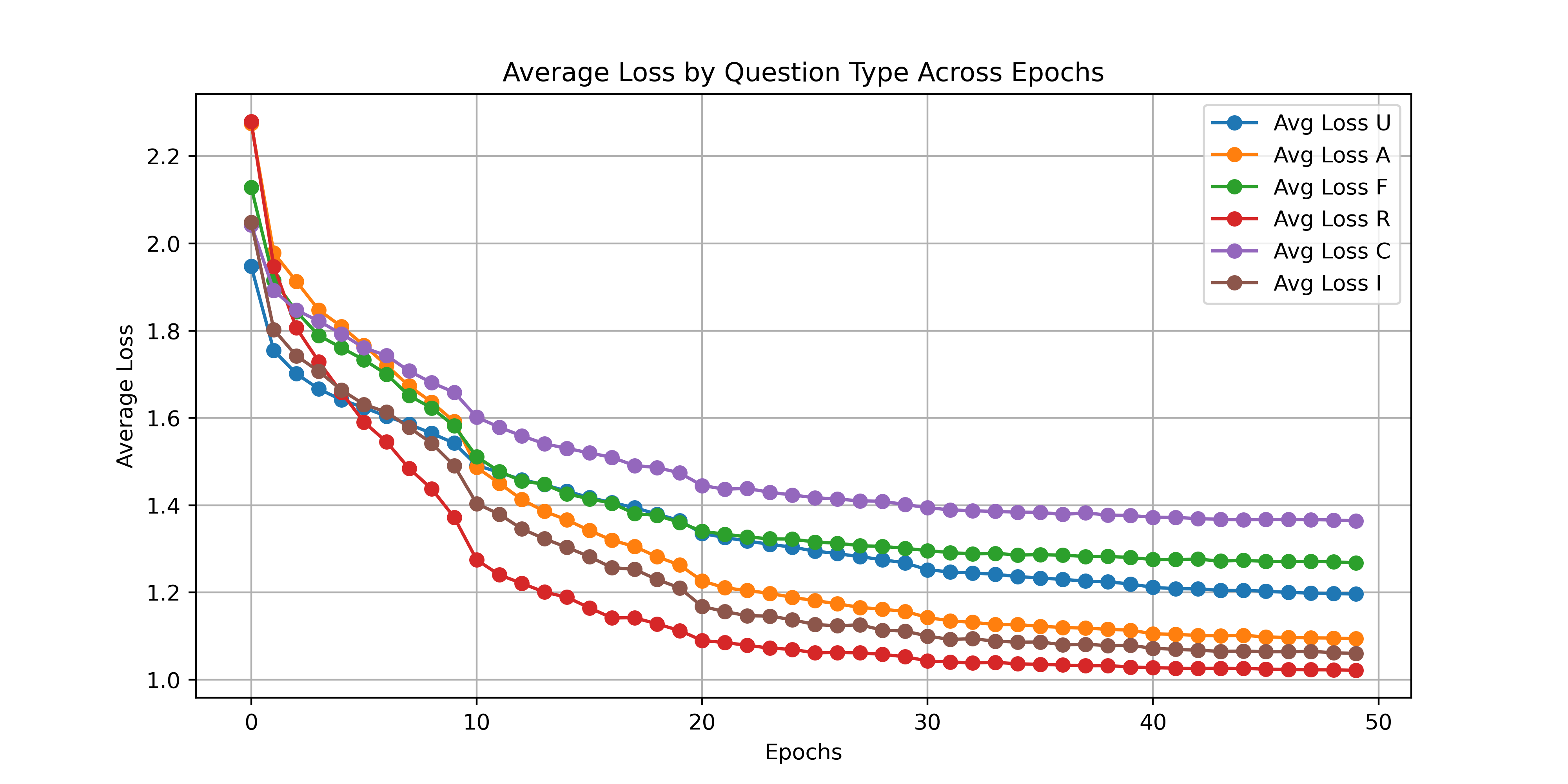}
    %\caption{}
    \label{fig:after_qtype_loss}
\end{subfigure}
\caption{Comparative visualization of training loss across different question types without (up) and with (down) the implementation of question-type guided attention.}
\label{fig:qtype_loss}
\end{figure}

In the ablation study section of this paper, we specifically and comprehensively explore the question-type guided attention mechanism, a core innovation in our model. Initially, as shown in Figure \ref{fig:qtype_loss}, we presents a comparative visualization of the loss across different question types, illustrating the effects before and after implementing the question-type guided attention mechanism. The graphs clearly demonstrate that, when the question-type guided attention is not used, there is an initial decline in loss across all types, followed by a subsequent sudden increase. In contrast, upon implementing the question-type guided attention mechanism, the loss across all question types decreases with ongoing training. 

This indicates that without targeted attention guidance, initial rapid learning may focus on the easiest-to-learn features, but as optimization of these features saturates, more challenging questions do not receive adequate attention and resources, leading to overfitting on simpler types and a subsequent degradation in learning and optimization performance. The specific performance metrics are presented in Table \ref{tab:attenabl}.

Additionally, to further demonstrate the impact of the question-type guided attention mechanism on the model’s performance across various question types, we applied this mechanism to advanced models individually, with the results presented in Table \ref{tab:advanced_qtg}.

\begin{table}[!t]
\centering
\begin{tabular}{@{}lcc@{}}
\toprule
w/o Q-Type Guided Attention  & acc (\%) & Orinial (\%) \\ 
\midrule
Basic Understanding        & 44.6     & 44.7 \\
Event Forecasting          & 39.5     & 44.8 \\
Reverse Reasoning          & 48.2     & 48.2 \\
Counterfactual Inference   & 51.9     & 56.8 \\
Introspection              & 39.2     & 41.9 \\
Attribute Attribution      & 46.8     & 47.8 \\ 
\midrule
Overall Avg-acc (\%)       & 45.3     & 45.6 \\
Overall IFWAA (\%)         & 44.3     & 47.6 \\
Overall EWAA (\%)          & 45.0     & 47.4 \\ 
\bottomrule 
\end{tabular}
\caption{Question-Type Guided Attention Ablation}
\label{tab:attenabl}
\end{table}

\begin{table*}[!t]
    \centering
    \begin{tabular}{@{}lccccccccc@{}}
        \toprule[0.15em]
        \multirow{2}{*}{\textbf{Setting}} & \multicolumn{9}{c}{\textbf{SUTD-TrafficQA}} \\
        \cmidrule(l){2-10}
        & \textbf{B} & \textbf{F} & \textbf{R} & \textbf{C} & \textbf{I} & \textbf{A} & \textbf{Avg-acc} & \textbf{IFWAA} & \textbf{EWAA} \\
        \midrule[0.15em]

        CMCIR \cite{liu2023crossmodalcausal} & 36.1 & 48.8 & \best{52.2} & 46.0 & 38.4 & \best{52.6} & 38.6 & 47.8 & 45.7 \\
        CMCIR $+$ \textbf{QTG Attn} & \best{36.4} & \best{50.2} & 51.2 & \best{48.3} & \best{41.7} & 51.5 & \best{39.7} & \best{48.9} & \best{46.6} \\
        \midrule
        
        Efficient-VideoQA \cite{lyu2023video} & 39.1 & 30.2 & 44.8 & 45.1 & 37.6 & \best{41.3} & 39.7 & 39.2 & 39.7 \\
        Efficient-VideoQA $+$ \textbf{QTG Attn} & \best{39.3} & \best{35.7} & \best{46.4} & \best{45.3} & \best{39.2} & 39.2 & \best{40.9} & \best{41.2} & \best{40.9} \\
        \midrule

        DualVGR \cite{9488296} & 33.9 & 41.6 & 50.6 & 41.4 & 33.4 & 50.6 & 36.1 & 40.5 & 41.9\\
        DualVGR $+$ \textbf{QTG Attn} & \best{34.4} & \best{46.2} & \best{51.7} & \best{41.8} & \best{33.7} & \best{50.9} & \best{40.3} & \best{41.9} & \best{43.1} \\
        \midrule
        
        \bottomrule[0.15em]
    \end{tabular}{}
    \caption{\textbf{QTG Attn} (Question-Type Guided Attention) as a plug-and-play method can help multiple models achieve better results. The best result on each module in each column is highlighted in \best{bold}.}
    \label{tab:advanced_qtg}
\end{table*}

\subsection{Temporal Autoregression Ablation}
To assess the impact of the Temporal Autoregression module on model performance, we designed ablation experiments. By removing the Temporal Autoregression module, we compared the differences in model performance on video question answering tasks before and after the modification. The results of the ablation experiment on the Temporal Autoregression module are presented in the Table \ref{tab:temabl}.

Analyzing the metrics results, it is evident that the accuracy across all question types decreases in the absence of the Temporal Autoregression (TA) module, with a more pronounced decline in question types that heavily depend on temporal sequences. For instance, the accuracy for ``Event Forecasting” questions drops from 44.8\% to 38.2\%, and for ``Attribute Attribution” questions, it decreases from 47.8\% to 43.4\%. The results indicate that the Temporal Autoregression module provides significant benefits to the model in assisting with temporal modeling and causal reasoning.

\begin{table}[!t]
\centering
\begin{tabular}{@{}lcc@{}}
\toprule
w/o TA model              & acc (\%) & Orinial (\%) \\ 
\midrule
Basic Understanding        & 43.4     & 44.7 \\
Event Forecasting          & 38.2     & 44.8 \\
Reverse Reasoning          & 46.5     & 48.2 \\
Counterfactual Inference   & 56.1     & 56.8 \\
Introspection              & 40.4     & 41.9 \\
Attribute Attribution      & 43.4     & 47.8 \\ 
\midrule
Overall Avg-acc (\%)       & 43.6     & 45.6 \\
Overall IFWAA (\%)           & 45.1     & 47.6 \\
Overall EWAA (\%)      & 44.7     & 47.4 \\ 
\bottomrule
\end{tabular}
\caption{Temporal Autoregression Ablation}
\label{tab:temabl}
\end{table}

\begin{figure}[!t]
    \centering
    \includegraphics[width=\linewidth]{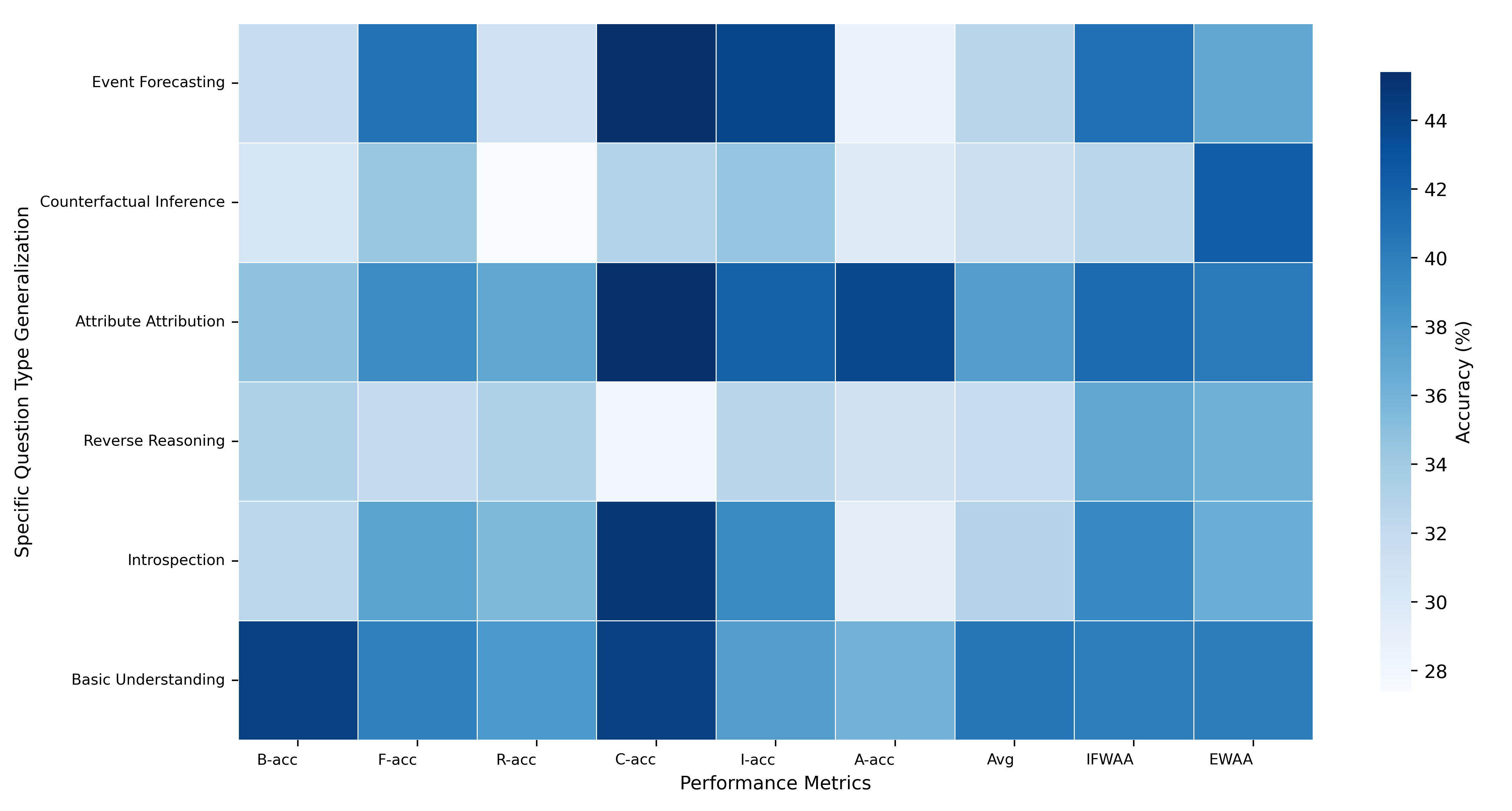}
    \caption{Generalization Results Heatmap.}
    \label{fig:heatmap}
\end{figure}

\subsection{Generalization Across Question Types}

In order to assess the impact of question type diversity on Video Question Answering (VideoQA) system performance, we conducted generalization experiments to test how well a model trained on one specific question type performs on unseen types. All available samples were used for training, and 25\% of each question type’s training set samples were allocated for validation. If a question type lacked sufficient samples for validation, all available samples were used.

\begin{table}[!t]
\centering
\begin{tabular}{@{}>{\centering\arraybackslash}p{1cm}ccccccc@{}}
\toprule
\multicolumn{1}{c}{} & \textbf{F-gl} & \textbf{C-gl} & \textbf{A-gl} & \textbf{R-gl} & \textbf{I-gl} & \textbf{U-gl} \\
\midrule
B-acc       & 31.7 & 30.3 & 34.8 & 33.3 & 32.5 & 44.3 \\
F-acc       & 40.8 & 34.4 & 39.0 & 32.0 & 37.2 & 39.9 \\
R-acc       & 31.1 & 27.4 & 37.1 & 33.2 & 35.5 & 38.1 \\
C-acc       & 45.4 & 33.0 & 45.4 & 28.0 & 44.9 & 44.3 \\
I-acc       & 43.9 & 34.6 & 41.9 & 32.7 & 39.2 & 37.8 \\
A-acc       & 28.7 & 29.7 & 43.8 & 31.0 & 29.3 & 36.1 \\
\midrule
Avg      & 32.7 & 31.4 & 37.7 & 31.7 & 32.9 & 40.6 \\
IFWAA    & 41.0 & 32.6 & 41.4 & 37.1 & 39.3 & 40.0 \\
EWAA     & 36.9 & 42.2 & 40.3 & 36.2 & 36.4 & 40.1 \\
\bottomrule
\end{tabular}
\caption{Generalization results. Rows indicate performance metrics for each question type, columns represent models trained on a specific question type.}
\label{tab:gener}
\end{table}

The generalization experiment results in Table \ref{tab:gener} and Figure \ref{fig:heatmap} show that the model trained on “Basic Understanding” question types performs the best across all question types, with an average accuracy of 40.6\%. This is likely because ``Basic Understanding” question types cover a wide range of basic scenarios, allowing the model to learn various general features and foundational patterns during training. For further analysis, please see the Appendix.

\section{Conclusion}
This paper introduced the QTG-VQA model, which improved performance in complex video question answering by using question-type-guided learning. Ablation studies confirmed the Temporal Autoregression module’s importance for tasks requiring temporal reasoning. The new evaluation metrics offered a more precise understanding of the model’s performance across question types. This study confirms the critical role of question type diversity in video question answering systems and highlights the generalization capabilities across different question types, offering valuable insights for future research in this field.

\bibliography{main}

\newpage
\appendix

\setcounter{secnumdepth}{2} % if section numbers are desired.

\section{Statistical details of VideoQA datasets}
In this section, we provide a detailed statistical overview of the VideoQA datasets mentioned in this paper, specifically focusing on those datasets that utilize a multi-choice task format. The statistics include various dimensions such as: dataset names and their publication details, whether question types are differentiated in the annotation files, data sources, the number of videos and questions, and whether annotations are created manually or automatically. 
For further details, please refer to Table \ref{tab:dataset}.

\begin{table*}[htbp]
\centering
\setcounter{table}{0}
\renewcommand{\thetable}{A.\arabic{table}} % Change table numbering for appendix
\begin{tabularx}{\textwidth}{@{}>{\hsize=1.6\hsize}X>{\hsize=0.4\hsize}X>{\hsize=0.8\hsize}X>{\hsize=0.6\hsize}X>{\hsize=0.4\hsize}X@{}} 
\toprule
Dataset & Q/type & Data Source & \#Video/\#QA & Annotation \\
\midrule
MovieQA \cite{tapaswi2016movieqa} & F & Movies & 6.7K/6.4K & Man \\
VideoQA(FiB) \cite{zhu2017uncovering} & T   & Multiple source  & 109K/390K & Auto \\
YouTube2Text-QA \cite{zhao2017video}  & F   & Web videos     & 1.9K/48K  & Auto  \\
PororoQA \cite{Kim2017DeepStoryVS} & T & Cartoon & 171/8.9K & Man \\
TGIF-QA \cite{jang2017tgif}  & T  & Animated GIF & 71K/165K & Auto, Man \\
EgoVQA \cite{fan2019egovqa} & T & Egocentric videos & 520/580 & Man \\
TVQA \cite{lei2019tvqa} & F & TV shows & 21K/152K & Man \\
TVQA+ \cite{lei2019tvqaplus} & F & TV shows & 4.1K/29K & Man \\
Social-IQ \cite{zadeh2019social} & F & Web videos & 1.2K/7.5K & Man \\
CLEVRER \cite{yi2019clevrer} & T & Synthetic videos & 10K/305K & Auto \\
LifeQA \cite{castro-etal-2020-lifeqa} & T & Web videos & 275/2.3K & Man \\
How2QA \cite{li2020hero} & F & Web videos & 22K/44K & Man \\
DramaQA \cite{choi2020dramaqa} & F & TV shows & 23K/17K & Man \\
PsTuts-VQA \cite{zhao2020video} & F & Tutorial videos & 76/17K & Man \\
KnowIT VQA \cite{garcia2020knowit} & $\mathrm{T^{-}}$ & TV shows & 12K/24K & Man \\
KnowIT-X VQA \cite{wu2021transferring} & $\mathrm{T^{-}}$ & TV shows & 12K/21K & Man \\
TGIF-QA-R \cite{peng2021progressive} & T & Animated GIF & 71K/165K & Auto \\
SUTD-TrafficQA \cite{xu2021sutd} & T & Traffic scenes & 10K/62K & Man \\
NExT-QA \cite{xiao2021nextqanext} & T & Web videos & 5.4K/52K & Man \\
NEWSKVQA \cite{gupta2022newskvqa} & T  & News videos & 12K/1M & Auto \\
Causal-VidQA \cite{li2022representation} & T & Web videos & 26K/107K & Man \\
NewsVideoQA \cite{jahagirdar2023watchingnewsvideo} & F & News viedos & 3k/8.6k & Man \\
STAR \cite{wu2024star} & T & Homemade videos & 22K/60K & Auto \\
Sports-QA \cite{li2024sports} & T & Sports videos & 6k/94k & Man \\
\bottomrule
\end{tabularx}
\caption{Summary of VideoQA Datasets. In the Q/type column, $\mathrm{T}$ indicates that the dataset includes question types, $\mathrm{F}$ indicates that the dataset does not include question types, and $\mathrm{T^{-}}$ means that question types are only available in the test set.}
\label{tab:dataset}
\end{table*}

\begin{figure*}[!h]
    \centering
    \setcounter{figure}{0}
    \renewcommand{\thefigure}{E.\arabic{figure}}
    \includegraphics[width=0.95\linewidth]{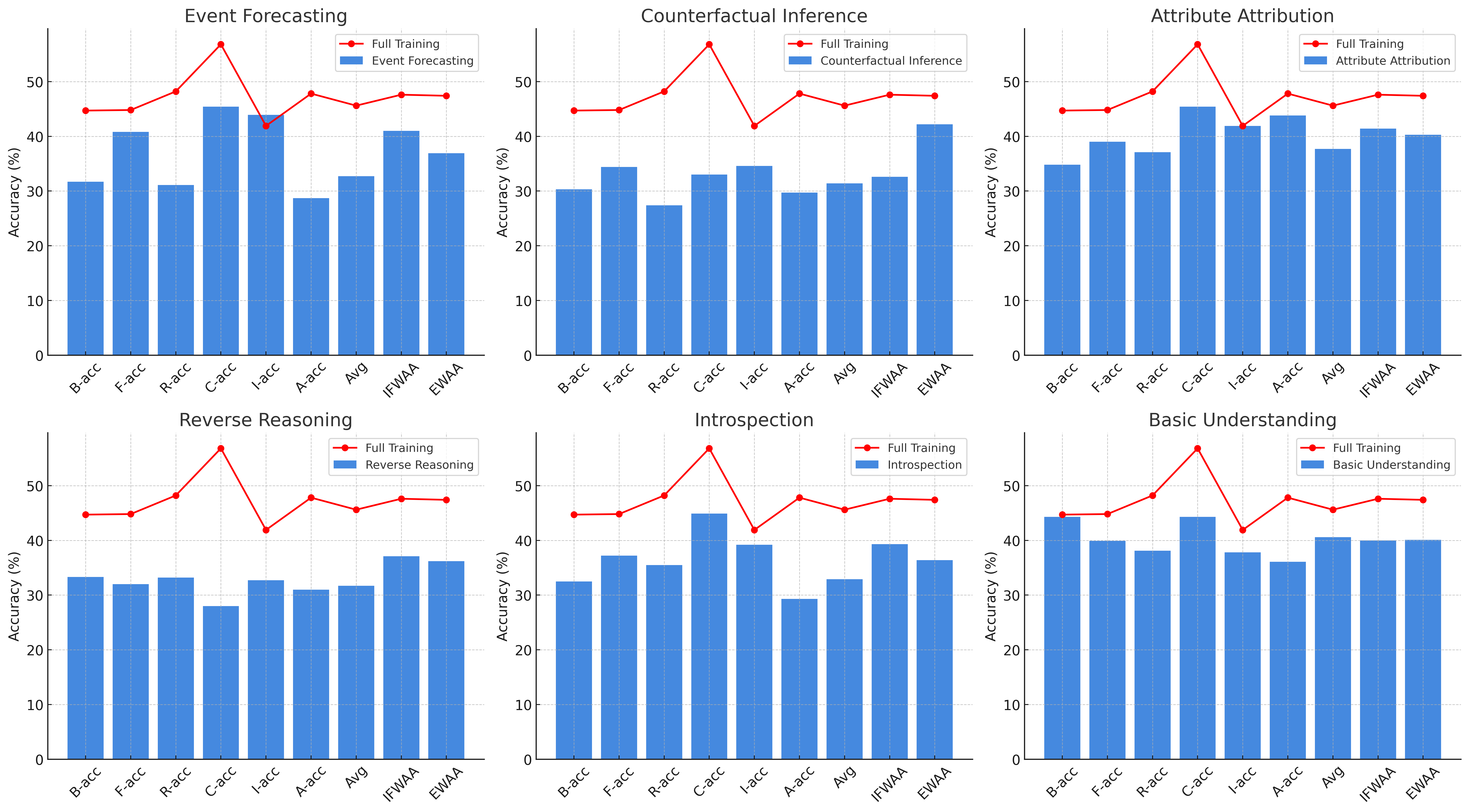}
    \caption{Generalization Performance Across Different Question Types. The bar chart for each question type displays the various metrics from the generalization training, while the red line represents the corresponding metrics from full training as a baseline.}
    \label{fig:genalone}
\end{figure*}

\section{Examples of SUTD-TrafficQA Question-Answer Pairs}
This section presents examples of question-answer pairs for various question types. Following the approach by \cite{chen2023tem}, we convert the question-answer pair texts into declarative sentences using specific linguistic templates. For detailed examples, please refer to Table \ref{tab:sutdexample}.

\begin{table*}[!t]
\renewcommand{\thetable}{B.\arabic{table}} 
\setcounter{table}{0} % Reset table counter to start with B.1
\centering
%\begin{tabular}{@{}lp{12cm}@{}}
\begin{tabularx}{\textwidth}{@{}lX@{}} 
%\toprule
\hline \hline
\addlinespace[3pt] 
\textbf{Question type} & \textbf{Basic Understanding} \\
\midrule
\textbf{Question} & What is the surrounding environment like in the video? \\
\textbf{Answers} & tunnel, mountainous, modern city, highway or expressway \\
\textbf{Declarative Sentences} & The surrounding environment like in the video is tunnel. The surrounding environment like in the video is mountainous. The surrounding environment like in the video is modern city. The surrounding environment like in the video is highway or expressway. \\
\hline \hline
\addlinespace[3pt]
\textbf{Question type} & \textbf{Attribution} \\
\midrule
\textbf{Question} & What could possibly cause this accident? \\
\textbf{Answers} & Obstructed by unexpected objects, Sudden braking of a vehicle, Violation of traffic rules by pedestrians, Sudden or extreme movement by a vehicle \\
\textbf{Declarative Sentences} & Obstructed by unexpected objects could possibly cause this accident. Sudden braking of a vehicle could possibly cause this accident. Violation of traffic rules by pedestrians could possibly cause this accident. Sudden or extreme movement by a vehicle could possibly cause this accident. \\
\hline \hline
\addlinespace[3pt]
\textbf{Question type} & \textbf{Introspection} \\
\midrule
\textbf{Question} & Could the accident be prevented if the roads are marked clearly? \\
\textbf{Answers} & Yes, the vehicles would stay in their lane. No, that was not the main cause of the accident. \\
\textbf{Declarative Sentences} & The accident could not be prevented if the roads are marked clearly, that was not the main cause of the accident. The accident could be prevented if the roads are marked clearly, the vehicles would stay in their lane. \\
\hline \hline
\addlinespace[3pt]
\textbf{Question type} & \textbf{Counterfactual Inference} \\
\midrule
\textbf{Question} & Will an accident happen if the current traffic load is tripled? \\
\textbf{Answers} & Yes, the road is too small. No, road is big enough.\\
\textbf{Declarative Sentences} & An accident happen if the current traffic load is tripled, the road is too small. An accident happen if the current traffic load is tripled, road is big enough. \\
\hline \hline
\addlinespace[3pt]
\textbf{Question type} & \textbf{Reverse Reasoning} \\
\midrule
\textbf{Question} & Which could be the reason for this accident? \\
\textbf{Answers} & Traffic light violation, Retrograde vehicles, Improper lane change, Obstructed view or limited visibility \\
\textbf{Declarative Sentences} & Traffic light violation could be the reason for this accident. Retrograde vehicles could be the reason for this accident. Improper lane change could be the reason for this accident. Obstructed view or limited visibility could be the reason for this accident. \\
\hline \hline
\addlinespace[3pt]
\textbf{Question type} & \textbf{Event Forecasting} \\
\midrule
\textbf{Question} & Are there any vehicles about to turn? \\
\textbf{Answers} & some, not any \\
\textbf{Declarative Sentences} & There are some vehicles about to turn, the vehicle is signalling a turn. There are not any vehicles about to turn. \\
%\bottomrule
\hline \hline
%\end{tabular}
\end{tabularx}
\caption{QA pairs example for each question type}
\label{tab:sutdexample}
\end{table*}

\section{Details of Baseline Models}
To explore common approaches to applying vision-language pre-trained models to video question answering tasks, Section 4 of this paper presents a comparative analysis of eleven baseline methods in contrast to our proposed QTG-VQA approach. In this section, we provide an overview of these baseline methods.
\begin{itemize}
    \item \textbf{Unsupervised CLIP} \cite{radford2021learning}. The CLIP model employs contrastive learning on a large scale of image-text pairs, acquiring rich joint visual-textual representations without the need for task-specific supervision. This enables it to handle complex visual question answering tasks even in the absence of explicit annotations.
    \item \textbf{Unsupervised CLIP + Language template}. Building upon the unsupervised CLIP model, predefined templates are used to transform QA pairs into declarative sentences, thereby reducing the semantic space gap between pre-training and downstream domains.
    \item \textbf{Totally Fine-tuning}. Retraining all parameters of the pre-trained model for downstream video question answering tasks can significantly enhance performance on specific tasks; however, this approach incurs substantial computational costs and is prone to overfitting.
    \item \textbf{Partially fine- tuning}.Partial fine-tuning techniques adjust only a subset of parameters in the pre-trained model. This paper specifically modifies the projection layer.
    \item \textbf{LORA} \cite{hu2021lora}. LORA applies low-rank adaptations to selectively modify model weights, allowing efficient fine-tuning with minimal additional parameters.
    \item \textbf{CLIP-Adapter} \cite{gao2024clip}. This approach integrates small, trainable adapter layers into the CLIP architecture to enhance task-specific performance without extensive retraining.
    \item \textbf{Multi-layer CLIP-Adapter}. Extends the CLIP-Adapter concept by inserting adapters at multiple layers, increasing the model's adaptability across different levels of representation.
    \item \textbf{Prompt learning (change words) with adapter heads} \cite{zhou2022learning}. Modifies the input prompts by changing words and utilizes adapter heads to fine-tune responses for better task alignment.
    \item \textbf{Prompt learning (change words) without adapter heads} \cite{zhou2022learning}. Similar to the above but relies solely on changing input prompts without using adapter heads, reducing complexity.
    \item \textbf{Prompt learning (add words)} \cite{jia2022visual}. Enhances prompt learning by adding words to the input, providing more contextual clues for the model.
    \item \textbf{Tem-adapter} \cite{chen2023tem}. By adding text adapters and visual adapters, the image-text model is transferred to the video domain to address downstream VQA tasks.
\end{itemize}

\begin{table*}[!t]
\centering

% Training Setup Table
\begin{tabular}{ccc}
\toprule
\multicolumn{3}{c}{\textbf{Training Setup}} \\
\midrule
initial learning rate & batch size & max training epochs \\
1e-4 & 128 & 50 \\
\bottomrule
\end{tabular}
\vspace{0.17cm} 

% SemanticAligner Parameters Table
\begin{tabular}{cccc}
\toprule
\multicolumn{4}{c}{\textbf{Main Module Parameters}} \\
\midrule
transformer width & number of heads & dropout rate & visual dimension \\
128 & 16 & 0.2 & 512 \\
encoder layer & decoder layer & qtype embedding layer & \phantom{empty} \\
1 & 1 & 1 & \\
\bottomrule
\end{tabular}
\vspace{0.17cm} 

% ContextDecoder Specific Parameters Table
\begin{tabular}{ccccc}
\toprule
\multicolumn{5}{c}{\textbf{ContextDecoder Fusion Module Parameters}} \\
\midrule
transformer width & number of heads & transformer layers & visual dimension & dropout rate  \\
256 & 4 & 6 & 512 & 0.1  \\
\bottomrule
\end{tabular}
\vspace{0.17cm}

% TempAligner Parameters Table
\begin{tabular}{ccccc}
\toprule
\multicolumn{5}{c}{\textbf{Temporal Autoregression Module Parameters}} \\
\midrule
module dimension & dropout rate & position-encoding length & encoder layer & decoder layer \\
512 & 0.1 & 5000 & 1 & 1 \\
\bottomrule
\end{tabular}
\renewcommand{\thetable}{D.\arabic{table}}
\setcounter{table}{0}
\caption{Experimental Hyperparameters}
\label{tab:hyperparameters}
\end{table*}

\section{Hyperparameter Settings}
This section outlines the detailed hyperparameter configurations employed in QTG-VQA model. To ensure the consistency and control of experimental variables, the hyperparameter settings described below in Table \ref{tab:hyperparameters} are uniformly adopted across all experiments. This standardization is maintained even in scenarios where modifying certain hyperparameters could potentially enhance performance. Such an approach ensures that the observed results are directly attributable to the experimental conditions specified, rather than variations in hyperparameter values.

\section{Detailed Analysis of Generalization Experiments}
In this section, we will provide a detailed analysis of generalization experiments across different question types, which aims to interpret the impact of question type diversity on the question-answering system. Firstly, we will analyze the overall generalization experiments. Figure \ref{fig:heatmap} presents a heatmap of the generalization results across all question types. This visualization illustrates the performance of the model when trained on specific question types and tested on unseen question types.

In the overall generalization experiments, the ``Basic Understanding'' question type showed the best generalization performance, achieving high accuracy across all question types. It also exhibited the most balanced and stable performance in terms of average accuracy (Avg-acc), Inverse Frequency Weighted Average Accuracy (IFWAA), and Equal Weight Average Accuracy (EWAA). ``Attribute Attribution'' followed closely, demonstrating strong cross-type learning capabilities. Notably, ``Basic Understanding'', ``Attribute Attribution'', ``Event Forecasting'', and ``Introspection'' types all generalized well to ``Counterfactual Inference'', indicating that the question type do not require specific training and can achieve high accuracy through generalization alone. Additionally, ``Event Forecasting'' excelled in generalizing to ``Introspection'' questions, suggesting that possessing some predictive ability aids the model’s performance in introspection-related tasks. In contrast, ``Counterfactual Inference'' and ``Reverse Reasoning'' showed weaker generalization, indicating that these question types have more independent characteristics and are difficult to generalize from other question types. This suggests that these types may require more targeted training strategies to improve model performance.

Furthermore, Figure \ref{fig:genalone} illustrates the model’s generalization performance across different question types compared to the full training (Full Training) results. Next, we will analyze the generalization results for each question type individually. 

\begin{itemize}
    \item \textbf{Basic Understanding}. The ``Basic Understanding'' question type performs closely to the full training results across all metrics, with similar accuracy observed across different question types. This indicates that models trained on this type of question have strong generalization ability, and the general features learned during training are stably applicable to other question types.
    \item \textbf{Attribute Attribution}. ``Attribute Attribution'' shows the second-best performance in generalization training, especially matching the full training results in ``Counterfactual Inference'' questions.
    \item \textbf{Event Forecasting}. ``Event Forecasting'' exhibits significant variation in generalization performance across different question types. It performs well below full training results in ``Basic Understanding'', ``Reverse Reasoning'', and ``Attribute Attribution'' but surpasses full training performance in other types, particularly in ``Introspection''.
    \item \textbf{Introspection}. ``Introspection'' also generalizes well overall, though it underperforms specifically in ``Attribute Attribution''.
    \item \textbf{Counterfactual Inference and Reverse Reasoning}. These two question types demonstrate limited generalization performance, with noticeable gaps compared to full training results across all metrics. This suggests that these question types are relatively independent, making it difficult for the model to generalize enough common features from other types.
\end{itemize}

\end{document}